%% file: main.tex
\documentclass[10pt,twocolumn,letterpaper]{article}

\usepackage{pkg/cvpr}
\usepackage{times}
\usepackage{epsfig}
\usepackage{graphicx}
\usepackage{amsmath}
\usepackage{amssymb}
\usepackage{multirow}
\usepackage{svg}
\usepackage{booktabs, arydshln}
\usepackage{multicol}
\usepackage{hhline}
\usepackage{enumitem}
\usepackage{float}

\usepackage[pagebackref=true,breaklinks=true,letterpaper=true,colorlinks,bookmarks=false]{hyperref}

\cvprfinalcopy 

\ifcvprfinal\fi
\begin{document}

\title{Hybrid Composition with IdleBlock: More Efficient Networks for Image Recognition}

\author{Bing Xu, Andrew Tulloch, Yunpeng Chen\thanks{Work was done during internship at Facebook}, Xiaomeng Yang, Lin Qiao\\
Facebook AI\\
}

\maketitle

\begin{abstract}
  \input{latex/1-abs.tex}
\end{abstract}

\section{Introduction \label{sec:intro}}
\input{latex/2-intro.tex}

\section{Idle \& IdleBlock: Motivation and Discussion \label{sec:idle}}
\input{latex/3-idle_block.tex}

\section{Hybrid Composition Networks \label{sec:hc}}
\input{latex/4-hc.tex}

\section{Experiments \label{sec:exp}}
\input{latex/5-exp.tex}

\section{Conclusion and Future work \label{sec:future}}
\input{latex/6-conclusion.tex}

\newpage

{\small
\bibliographystyle{latex/pkg/ieee_fullname}
\bibliography{egbib}
}

\end{document}

%% file: latex/1-abs.tex
We propose a new building block, IdleBlock, which naturally prunes connections within the block. To fully utilize the IdleBlock we break the tradition of monotonic design in state-of-the-art networks, and introducing hybrid composition with IdleBlock. We study hybrid composition on MobileNet v3 \cite{howard2019searching} and EfficientNet-B0 \cite{tan2019efficientnet}, two of the most efficient networks. Without any neural architecture search, the deeper “MobileNet v3” with hybrid composition design surpasses possibly all state-of-the-art image recognition network designed by human experts or neural architecture search algorithms. Similarly, the hybridized EfficientNet-B0 networks are more efficient than previous state-of-the-art networks with similar computation budgets. These results suggest a new simpler and more efficient direction for network design and neural architecture search.

%% file: latex/2-intro.tex
Convolutional Neural Networks (CNN) have dominated computer vision tasks in recent years. Since AlexNet \cite{krizhevsky2012imagenet}, the computer vision community has sought improved CNN designs to make the backbone network more powerful and more efficient. Remarkable single branch backbones include Network in Network \cite{lin2013network}, VGGNet \cite{simonyan2014very}, ResNet \cite{he2016deep}, DenseNet \cite{huang2017densely}, ResNext \cite{xie2017aggregated}, MobileNet v1/v2/v3 \cite{howard2017mobilenets, sandler2018mobilenetv2, howard2019searching}, and ShuffleNet v1/v2 \cite{zhang2018shufflenet, ma2018shufflenet}. In recent years, backbone networks with multiple resolutions have also attracted attention from the research community. To learn with multiple resolutions, researchers design complex connections inside a block to handle information exchanges of different resolutions. Some efficient examples of this approach include the MultiGrid-Conv \cite{ke2017multigrid}, OctaveConv \cite{chen2019drop}, and HRNet \cite{sun2019deep}. All these contributions have significantly developed the philosophy of backbone network design.


To design more efficient CNNs, there are two mainstream efforts: Neural Architecture Search (NAS) and Network Pruning (NP). The motivation of neural architecture search is: given a constraint on computation resources, NAS attempts to automatically determine the best network connections, block designs and hyper-parameters. Hyper-parameter searching is a classic topic in machine learning, and in our paper we refer NAS as specially to searching over the connections and block design of neural network. The motivation of network pruning is: given a pretrained network, an automatic algorithm is able to remove unimportant connections, to reduce computation and parameters. 

In contrast to NAS over connections and NP, EfficientNet \cite{tan2019efficientnet} provides a joint hyperparameter for a backbone: the depth scaling factor $d$, width scaling factor $w$ and input resolution scaling factor  $r$, which is referred as compound scaling factors. Based on a variant of MobileNet v3, these jointly-searched scaling factors make the EfficientNet family 5$\times$ to 10$\times$ more efficient than all previous backbones in measured of computation cost (MAdds) or the number of parameters. 




\begin{figure}
    \centering
    \includegraphics[scale=0.53]{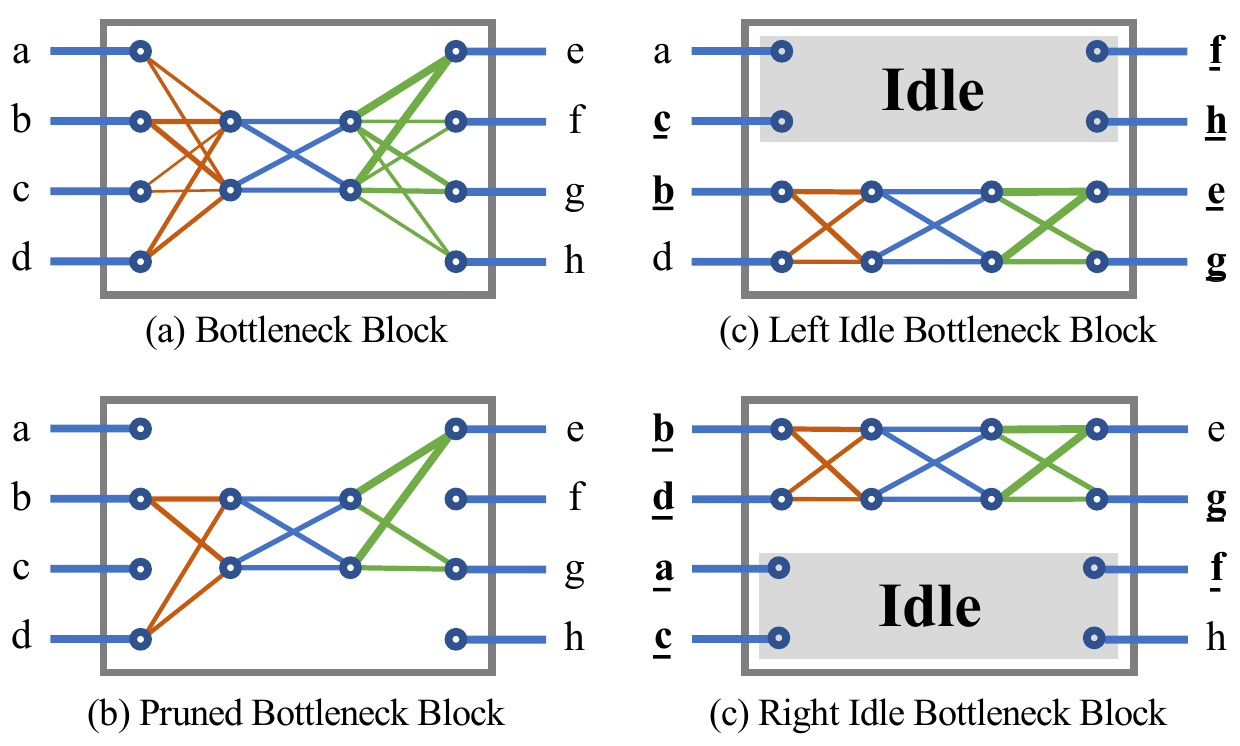}
    \caption{The motivation of our proposed Idle design. In Idle design information exchange is applied outside of the Idled blocks.}
    \label{fig:motivation}
\end{figure}

\begin{figure}
    \centering
    \includegraphics[scale=0.53]{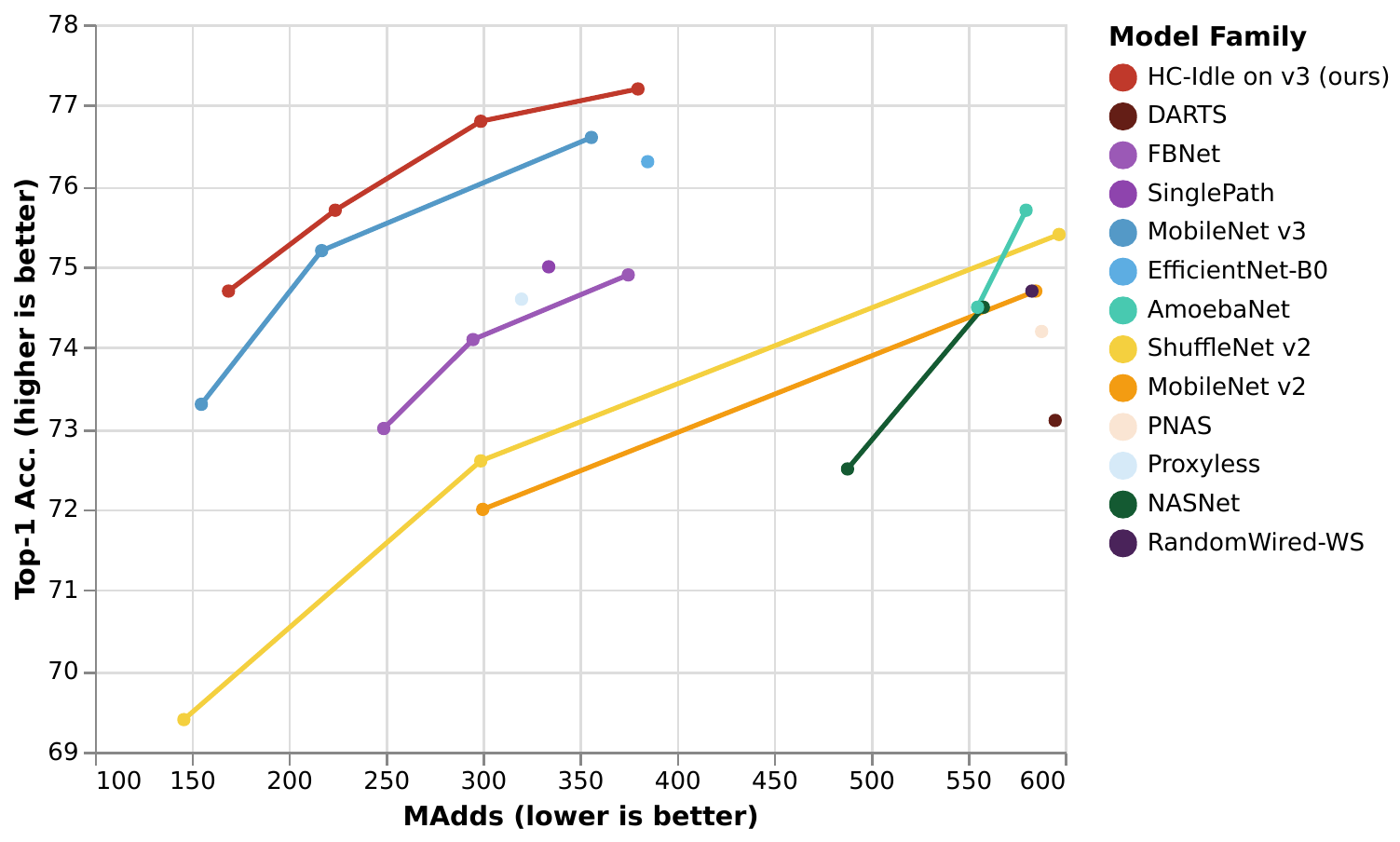}
    \caption{ImageNet Top-1 Accuracy / Computation cost trade-off for small model (MAdds $<$ 600M). In this figure we compare Hybrid Composition with IdleBlock on MobileNet v3 with state-of-art human expert-designed networks and NAS networks.}
    \label{fig:my_label}
\end{figure}


In this work, We revisit the modern workflow of obtaining efficient convolutional networks, and partition the problem into two separate steps. In the first step, we design a network architecture, and in the second step, we prune connections from the network. 

\begin{figure}
  \centering
  \includegraphics[scale=0.5]{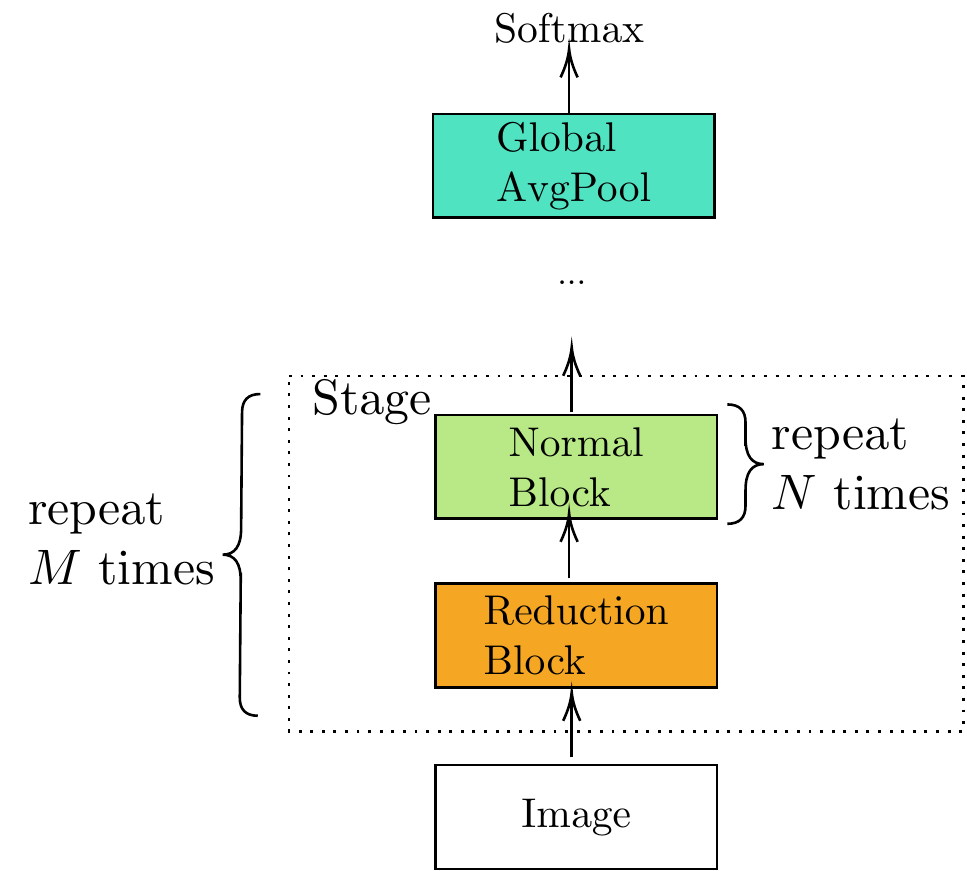}
  \caption{Monotonous design.}
  \label{fig:mono}
\end{figure}

In the first step, we note a common pattern in both human expert-designed architectures and searched architectures: for each backbone, the architecture is characterised by the design of a normal block and a reduction block. At the beginning of each stage, we insert a reduction block, and repeatedly stack the normal block. We repeat each stage multiple times, and for each stage we may have different number of normal blocks. We call this design pattern Monotonous Design (Figure \ref{fig:mono}). For example, ResNet monotonically repeats a Bottleneck block, ShuffleNet monotonically repeats a ShuffleBlock, MobileNet v2/v3 and EfficientNet monotonically repeats and Inverted Residual Block (MBBlock), NASNet repeats a Normal Cell, and FBNet repeats a variant of MBBlock with different hyper-parameters. For all the state-of-art networks, to our best knowledge, the blocks are guaranteed to have full information exchange. In the second step, some connections are pruned, and it is not be guaranteed that each block has full information exchange.

\begin{figure}
  \centering
  \includegraphics[scale=0.5]{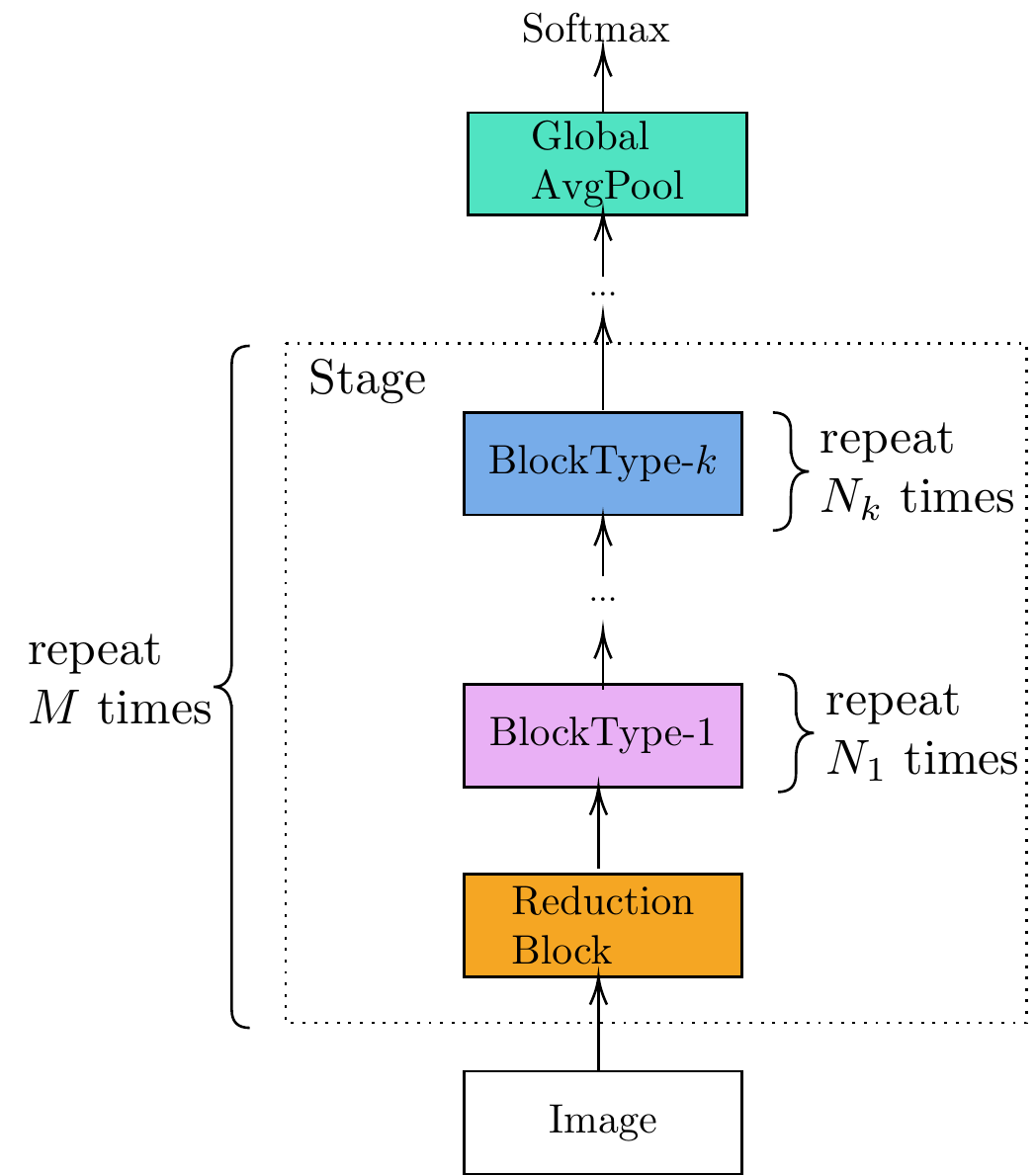}
  \caption{Hybrid Composition}
  \label{fig:hc}
\end{figure}


In this paper, we seek to design a more efficient network for image recognition tasks, by considering pruning in the network design step. We create a new block design methodology: \textit{Idle}. In the Idle design, a subspace of the input is not transformed: it is simply “idle” and passed directly to the output (Figure \ref{fig:my_label}). We also break the monotonous design constraint in state-of-the-art architectures. We call our non-monotonous composition method \textit{Hybrid Composition} (HC) (Figure \ref{fig:hc}).

The results are initially as expected --- if we monotonically compose a network with IdleBlock, we obtain a pruned network with acceptable accuracy loss. If we use hybrid composition with IdleBlock (and MBBlock), we significantly reduce the accuracy loss while substantially saving computation. The surprising result is this: \textbf{leveraging the pruned computation budget from IdleBlock to go deeper with hybrid composition achieves new state-of-art network architectures --- without complex multi-resolution design or neural architecture search.}

The paper is organized as follows. In Section \ref{sec:idle}, We discuss modern convolutional network block design, and introduce the Idle design and IdleBlock. In Section \ref{sec:hc}, we introduce Hybrid Composition (HC) with IdleBlock. In Section \ref{sec:exp} we study applying HC with IdleBlock to MobileNet v3 and EfficientNet-B0, which leads to new state-of-the-art network architectures under equivalent computation budgets. We also conduct ablation studies on our design choice. Section \ref{sec:future} contains conclusion and future work.

%% file: latex/3-idle_block.tex
In this section, we will revisit key milestones in convolution building block design. We will then introduce our Idle design and Idled version of MBBlock: IdleBlock.

\subsection{Bottleneck Block \& Inverted Residual Block}
\begin{figure}
    \centering
    \includegraphics[scale=0.23]{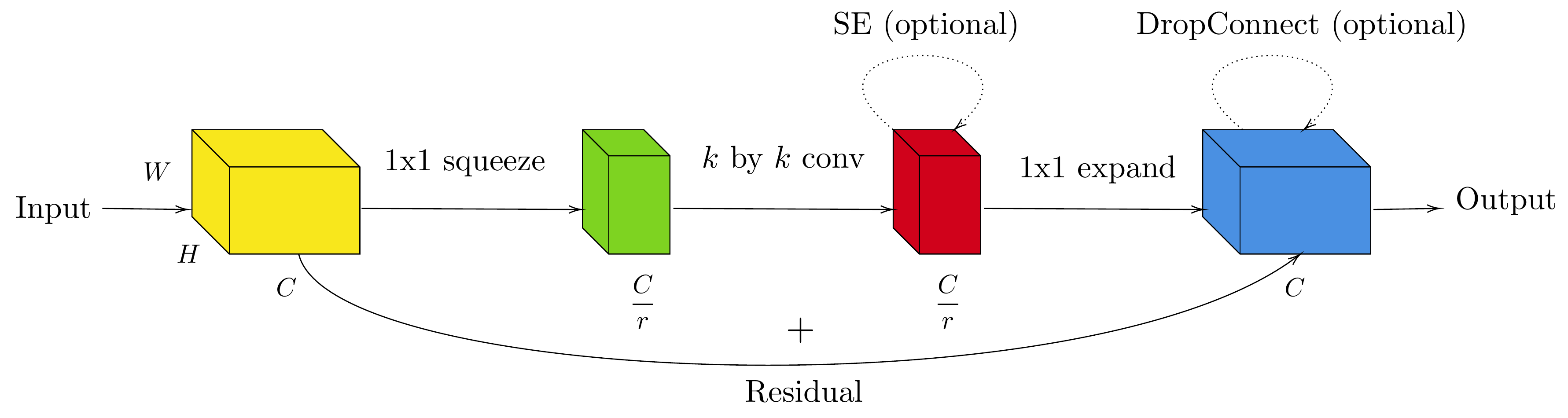}
    \caption{Bottleneck block. The Bottleneck block seeks to reduce computation of spatial convolution. Each block consists of expanded input and output, without nonlinearities. The residual connection is between expanded representations.}
    \label{fig:bottleneck}
\end{figure}

\begin{figure}
    \centering
    \includegraphics[scale=0.21]{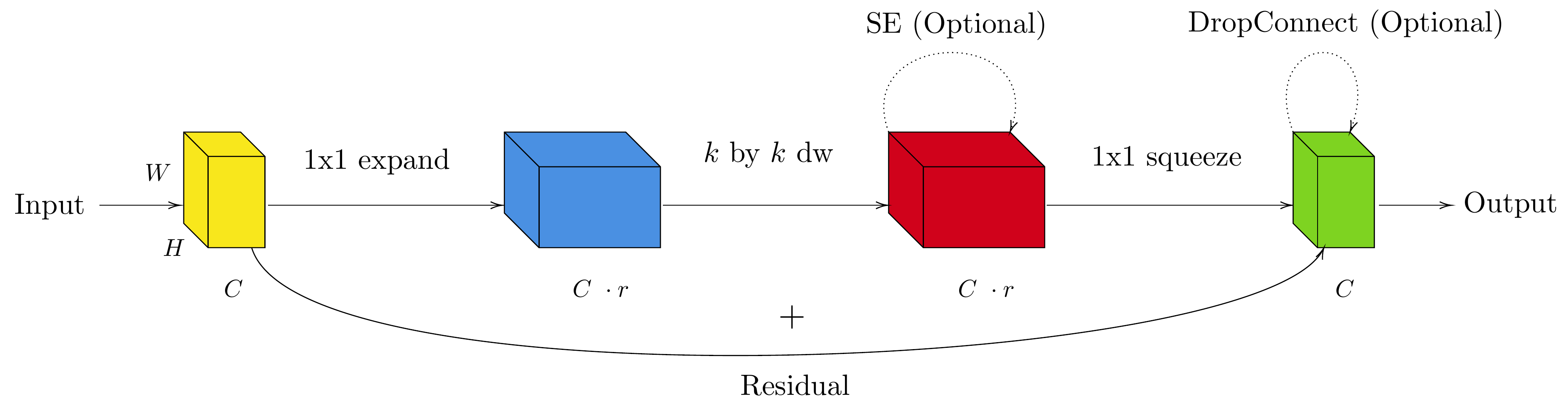}
    \caption{Inverted Residual Block (MBBlock). The Inverted Residual Block seeks to extract rich spatial information from an expanded projection. Each block consist of narrow inputs and outputs, without nonlinearities. The residual connection is between narrowed representations.}
    \label{fig:mbblock}
\end{figure}

The Inverted Residual Block (MBBlock) (Figure \ref{fig:mbblock}) is the fundamental building block behind success of MobileNet v2/v3 and EfficientNet. In MBBlock, a pointwise convolution will expand input dimension by a factor of $r$, apply a depthwise convolution to the expanded features, then apply another pointwise convolution to squeeze the high dimension features into a lower dimension (usually the same as input to allow residual addition). The key difference between the Bottleneck block \cite{he2016deep, howard2017mobilenets} design (Figure \ref{fig:bottleneck} and MBBlock is --- Bottleneck applies a spatial transformation ($k \times k$ convolution or $k \times k$ depthwise convolution) on a narrowed feature map, and MBBlock applies a spatial transformation on an expanded feature map. The Bottleneck design is proposed in ResNet, where $3 \times 3$ convolutions are the most computationally intensive component. With $3 \times 3$ convolutions, narrowed feature maps significantly reduce computation cost. However, when depthwise convolutions are used, the spatial transformation is no longer a computation bottleneck. A $k \times k$ depthwise convolution corresponds to running a linear regression with $k^2$ input elements repeatly in each channel. Intuitively, with more channels, the spatial transformation capacity will be higher. Empirically, with similar computational budgets (MAdds), MobileNet v1 in Bottleneck design achieves 70.6\% Top 1 accuracy on ImageNet \cite{deng2009imagenet}, while MobileNet v2 (1.4) achieves 74.7\% Top 1 accuracy \cite{sandler2018mobilenetv2}. The accuracy gap between MobileNet v1 and v2 indicates that for depthwise convolutions, an expanded feature map is helpful to improve a network's expression power. 


\subsection{ShuffleBlock v1/v2}

ShuffleBlock v1/v2 are depthwise convolution blocks that heavily rely on the channel shuffle operation --- a lightweight information exchange operation.

\begin{figure}
    \centering
    \includegraphics[scale=0.23]{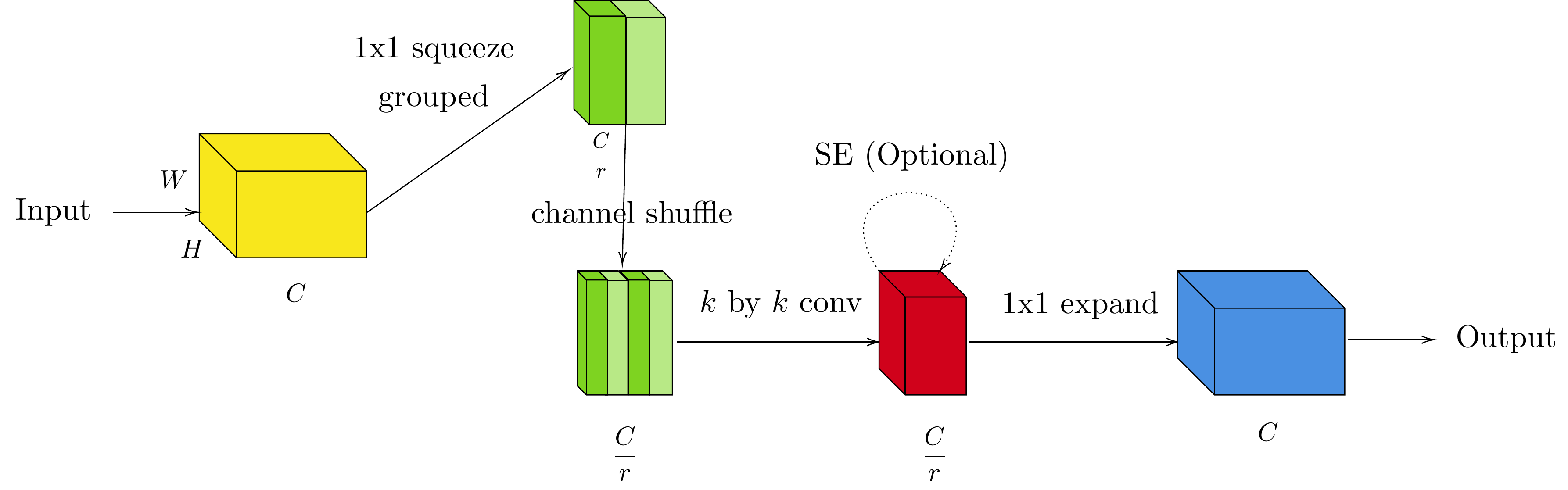}
    \caption{ShuffleBlock v1. ShuffleBlock v1 is an extension of Bottleneck block (Figure \ref{fig:bottleneck}). To reduce computation of the narrowed representation, a grouped pointwise operation is introduced, followed by channel shuffle operation.}
    \label{fig:shufflev1}
\end{figure}

ShuffleBlock v1 (Figure \ref{fig:shufflev1}) is inherited from Bottleneck block. A grouped pointwise convolution is used to replace the first pointwise convolution in Bottleneck block to narrow the channel dimension. A channel shuffle operator is followed by grouped pointwise convolution operator to provide information exchange between groups. The block topology is the same as the Bottleneck block.

\begin{figure}
    \centering
    \includegraphics[scale=0.25]{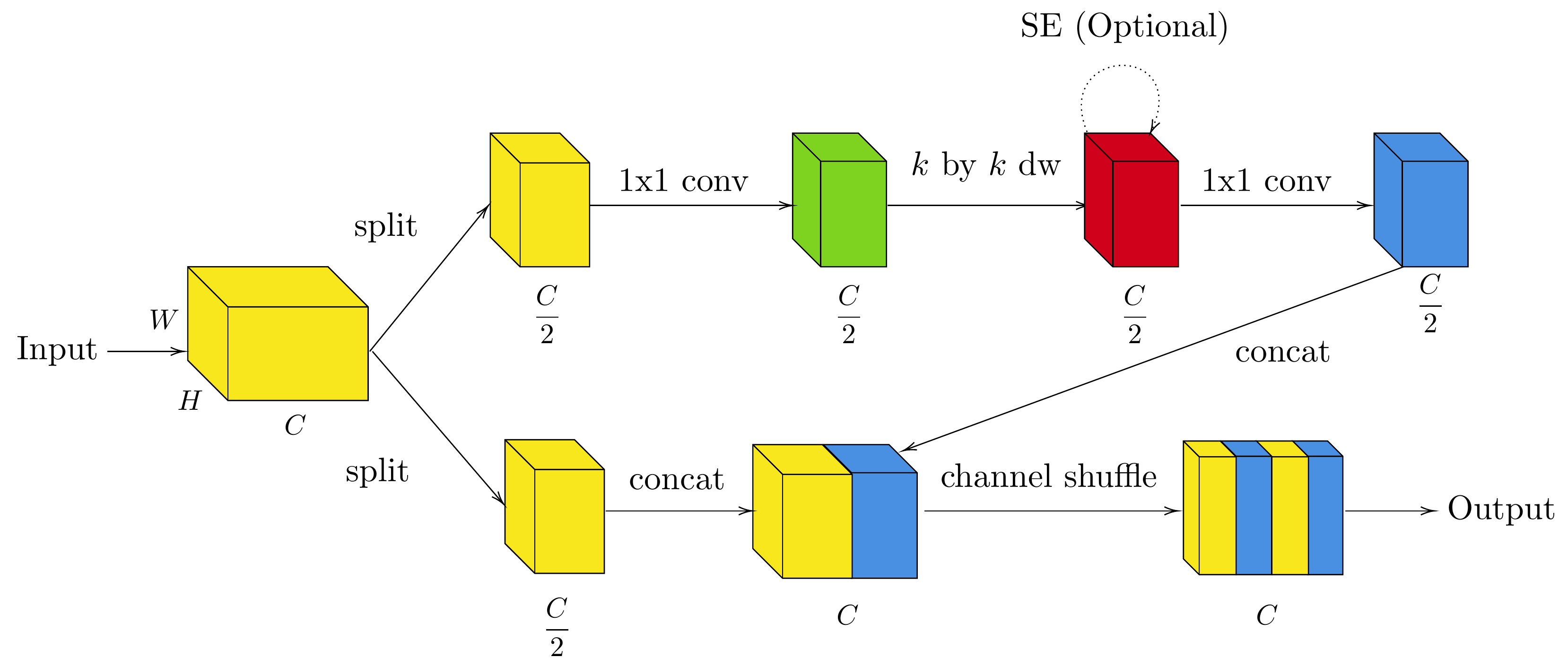}
    \caption{ShuffleBlock v2. ShuffleBlock v2 removed the grouped pointwise operation -- instead, using split to obtain a narrowed representation. Similar to Bottleneck block (Figure \ref{fig:bottleneck}) and ShuffleBlock v1 (Figure \ref{fig:shufflev1}), each block consist of an expanded input and output.}
    \label{fig:shufflev2}
\end{figure}

ShuffleBlock v2 (Figure \ref{fig:shufflev2}) is a following work of ShuffleBlock v1. Similar to Bottleneck block and ShuffleBlock v1, ShuffleBlock v2’s input and output are expanded feature maps. To further reduce computation cost, the group pointwise convolution which is used for narrowing feature map is removed. Instead, expanded feature map is split into two equal channel narrow feature maps: one narrowed feature map is transformed with a special Bottleneck block without internal dimension change,  the other feature map keeps the same until it is shuffled with transformed feature map. 

However, the channel shuffle operation is not efficiently realizable on many heterogeneous hardware, such as new neural network accelerators. For example on iPhone equipped with Apple Neural Engine (ANE) using CoreML, networks without channel shuffle operations run up to 10 times slower than equivalent networks without channel shuffles.

\subsection{"Idle" Design}

When we think about design principles for an efficient block structure, attempting to reduce computation (MAdds) is an obvious direction. For example, to reduce computation in a $k \times k$ spatial convolution, depthwise convolutions can be used to replace the original convolution operator (as in MBBlock). 

When we optimize directly for modern hardware platforms, there are more challenges than simply reducing MAdds. For example, post-training pruning will reduce computation, but without very specialized implementations and high levels of sparsity, it is hard to realize these computational improvements in practice. We seek to introduce a structured pruned topology into our block design, which allows us to achieve real-world speed ups.  With this insight, we create a new design pattern: \textit{Idle}, which aims to directly pass a subspace of the input to the output tensor, without any transformations. Figure \ref{fig:motivation} demonstrates motivation of Idle and network pruning.  In Idle design, we introduce an “idle” factor $\alpha \in (0, 1)$, which is also can be viewed as pruning factor. Given an input tensor $x$ with $C$ channels, the tensor will be sliced to two branches:  one “active” branch $x_1$ containing $C \cdot (1-\alpha)$ channels, and outputting a tensor $y_1$ with $C \cdot (1-\alpha)$ channels; and the other “idle” branch $x_2$ contains $C \cdot \alpha$ channels, copied directly to the output tensor $y$ with $C$ channels directly.


\subsubsection*{Contrasted with Residual Connections} With residual connections, the input tensor $x$ is added to the output tensor. In the Idle design, $C \cdot \alpha$ channels of input tensor $x$ will be copied into the output tensor directly, without any elementwise addition.

\subsubsection*{Contrasted with Dense Connections} In a densely connected block, the entire input $x$ is part of the output $y$, and entire input $x$ in used in the block transformation. In Idle design, only $C \cdot \alpha$ channels of $x$ are copied into the output tensor directly, and the other $C \cdot (1-\alpha)$ channels are used in the block transformation.

\subsubsection*{Contrasted with ShuffleBlock v2} The Idle design is closely related to ShuffleBlock v2 but with a few differences. First, ShuffleBlock v2 requires expanded feature map as input and outputs an expanded feature map; Idle is a design pattern doesn’t require on input/output dimension. Second, in ShuffleBlock v2, all information will be exchanged by using channel shuffle operator; In Idle design, we intend to don’t apply any change for a subspace of input. Third, Idle design’s motivation is to design a pre-pruned structure, ShuffleBlock v2 aims to obtain a narrowed feature map for spatial transformation. We can view ShuffleBlock v2 in a perspective of Idled Bottleneck block with extra channel shuffle.


\subsection{IdleBlock}

We summarize a few intuitive and empirical lessons we learned from ShuffleBlock v1/v2 and MBBlock:

\begin{enumerate}
    \item We need to apply the depthwise convolution on an expanded feature map. (MobileNet v1 vs MobileNet v2)
    \item Grouped convolutions are not necessary. (ShuffleNet v1 vs ShuffleNet v2)
    \item The channel shuffle operation is not friendly to various accelerators, and should be avoided.
\end{enumerate}

Based on these lessons, we introduce IdleBlock, an Idled version of MBBlock. 

\begin{figure}
    \centering
    \includegraphics[scale=0.22]{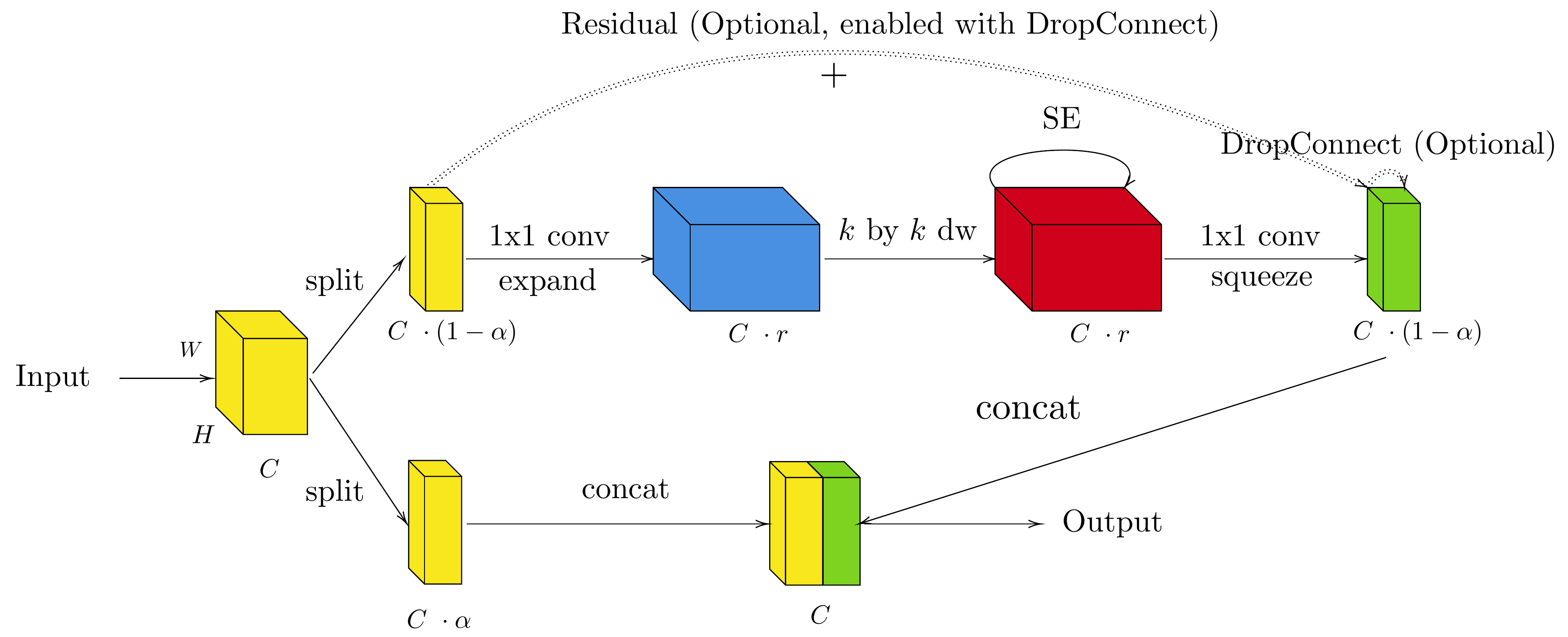}
    \caption{L-IdleBlock. IdleBlock is an idled version of MBBlock (Figure \ref{fig:mbblock}. Similar to MBBlock, each block consist of narrowed input and output without nonlinearities. L-IdleBlock concatenate non-transformed branch first, followed by transformed branch. R-IdleBlock concatenate in reversed way. L/R-IdleBlock won't change input/output shape or channel dimension.}
    \label{fig:L-idle}
\end{figure}

There are two variants of IdleBlock. If we use the concatenation function $\text{concat}(y_1, x_2)$ when constructing the output tensor from the two splits, we refer to the it as an L-IdleBlock (Figure \ref{fig:L-idle}), and if we use the concatenation function $\text{concat}(x_2, y_1)$, we refer to it as an R-IdleBlock. If an information exchange block is strictly followed by an IdleBlock, the L-IdleBlock and R-IdleBlock are equivalent. When stacking two or more IdleBlocks, a mix of L/R-IdleBlock units is different to a monotonic composition of L/R-IdleBlock units. 

A key distinction of mixed composition is the enhanced receptive field of the stacked output. Using two stacked IdleBlocks as example. Given inputs with receptive field $R_0$, if we monotonically stack two R-IdleBlocks, the first output branch’s receptive field will remain as $R_0$ as the first branch is idle, while the second output branch’s receptive field is $R_0 \cdot k^2$ as the second branch of input is updated by two $k$ by $k$ depthwise convolution operations. If we mix an L- and R-IdleBlock, L/R-IdleBlock, both of the output branches will be updated with one $k$ by $k$ depthwise convolution operator, which will change the receptive field of both output branches to $R_0 \cdot k$.


Given an MBBlock with input tensor shape $(1, C, H, W)$, expansion factor $r$,  stride $s$, and depthwise convolution kernel $k$, the theoretical computation complexity is

\begin{align}
 \sum (& 2 \cdot r \cdot C^2 \cdot HW, && \text{//1x1 expand} \nonumber \\ 
 & 2 \cdot r \cdot C \cdot (k^2 \cdot \frac{HW}{s^2}), && \text{//k by k depthwise}\ \\ 
 & 2 \cdot r \cdot C^2 \cdot \frac{HW}{s^2}) && \text{//1x1 squeeze} \nonumber 
\end{align}

For IdleBlock, the theoretical computation complexity is

\begin{align}
 \sum( &2 \cdot r \cdot C^2 \cdot (1-\alpha) \cdot HW, && \text{//1x1 expand} \nonumber \\ 
 & 2 \cdot r \cdot C \cdot (k^2 \cdot \frac{HW}{s^2}), && \text{//k by k depthwise}\ \\ 
 & 2 \cdot r \cdot C^2 \cdot (1-\alpha) \cdot \frac{HW}{s^2}) && \text{//1x1 squeeze} \nonumber
\end{align}

By replacing a MBBlock with an equivalent IdleBlock, we save: 

\begin{align}
    2 \cdot \alpha \cdot r \cdot C^2 \cdot (HW + \frac{HW}{s^2}) 
    \label{eq:save}
\end{align}




%% file: latex/4-hc.tex
In the previous section we introduced the IdleBlock. In this section we introduce Hybrid Composition (HC), a novel non-monotonic network composition method. 

In hybrid composition, at each stage of the network, we use a non-monotonic composition of multiple types building blocks. This is only possible if the different blocks have identical constraints on the input and output dimensions. 
For example, ShuffleBlock v1/v2 can be hybridized with a Bottleneck block, because both of the blocks accept a narrow input and output an expanded tensor. A Bottleneck block is not able to be hybridized with MBBlock, because these two blocks contains different bottleneck designs and the input and output dimensions are very different. Hybrid composition attempts to jointly utilize the different properties of multiple building blocks, which monotonically design is not able to do.

For a monotonic MBBlock network, such as MobileNet v2/v3 and EfficientNet, the first pointwise convolution operator in MBBlock is only used to expand the input dimension for the depthwise operation.

In our case with IdleBlock, both IdleBlock and MBBlock satisfy the input and output constraints for hybrid composition. Moreover, once we hybridize IdleBlock with MBBlock, the first pointwise convolution operator in MBBlock will be able to help us exchange information of two branches in IdleBlock, without an explicit channel shuffle operation as in ShuffleBlock.   

However, hybrid composition introduces another challenge. If a network stage contains $n$ MBBlock units, there are $2^n$ candidate combinations of MBBlock and IdleBlock placements in the Idle network, but we seek to explore only explore a small subset of these candidates.

To solve the challenge, we explore three configurations for the hybrid composition of MBBlock with IdleBlock: Maximum, None and Adjacent.

\subsubsection*{Maximum configuration}
We only use MBBlock as reduction block, or when the network changes the output dimension. All other MBBlocks will be replaced by IdleBlock. This is the computational lower bound of hybrid composition with IdleBlock, as we need at least one MBBlock unit to exchange information in each stage.

\subsubsection*{None configuration}
In this configuration, none of any blocks in the stage will be replaced with IdleBlock. This is the the computational upper bound of hybrid composition with IdleBlock. 

\subsubsection*{Adjacent configuration}
Adjacent configuration is a greedy strategy. We iteratively replace each MBBlock that has an MBBlock input with an IdleBlock, stopping when we reach our specified computational budget.


When we set $\alpha=0.5$ in the IdleBlock during the repeated stage (stride=1), we notice that according to Equation \ref{eq:save}, the theoretical computation cost of one MBBlock is roughly equal to cost of two IdleBlocks. This means that in the Adjacent configuration, to keep same computation complexity, we can replace one MBBlock with two IdleBlocks --- resulting in an approximately 30\% increase in overall network depth with same computation budget. We denote this special case as "Adjacent +1 IdleBlock" configuration.




Intuitively, a neural architecture search over the space of block placement policies should fall (in terms of accuracy and computation cost) between the lower bound and upper bounds described here (Maximum and None).

%% file: latex/5-exp.tex
We present experimental results on ImageNet 2012 classification dataset \cite{deng2009imagenet} to demonstrate the effectiveness of hybrid composition with IdleBlocks. We set pruning factor $\alpha$ to 0.5 for all our experiments. We focus on studying MobileNet v3 and EfficientNet-B0, two state-of-art efficient image recognition architectures. We report execution times on Intel Xeon CPUs and ARM Cortex CPUs. We also conduct ablation studies to validate our design decisions. In this section, we use "Top-1" to refer "Top-1 Accuracy" on ImageNet.

\subsection{Training Setup}

We use MXNet \cite{chen2015mxnet} and GluonCV \cite{guo2019gluoncv} for both single node training and distributed training. Our source code and pretrained models can be found at: URL.

\subsubsection*{Single Node}
We provide MobileNet v3 experiment results on single node with 8 NVIDIA V100 GPU. We use fp16/32 mixed-precision during training. The batch size is set to 256 for each GPU (2048 in total). We use a Nesterov optimizer with learning rate set to 2.6 and weight decay set to 3.0e-5 for all models. All models are trained with 360 epoches. We warm up training in the first 5 epochs.

\subsection*{Distributed}
We provide both MobileNet v3 and EfficientNet-B0 experiment results in distributed setting. All models are trained with 8 nodes, contains 64 V100 in total. Without special note, batch size is set to 128 per GPU.  We use LBSGD \cite{lin2019dynamic} to optimize all models, with initial learning rate 0.8, momentum 0.9, L2 weight decay 3.0e-5. All models are trained with 80 epochs (equivalent to 640 epoches on single node). We warm up training in the first 5 epoches. All distributed models are trained with Mixup \cite{zhang2017mixup} augmentation.


\subsection{Inference Setup}

We evaluate the inference time on Google Cloud Engine in single threaded and multi-threaded (4 threads) settings, on an Intel SkyLake CPU @ 2.0GHz. For ARM Cortex CPUs, we evaluate on an ARM Cortex-A53 @ 600MHz. All models are compiled with TVM \cite{chen2018tvm} and AutoTVM \cite{chen2018learning}, which provides substantial latency improvements on these architectures.

\subsection{Result on MobileNet v3}

\input{latex/tab/tab_mbv3.tex}
\input{latex/tab/tab_mbv3_all.tex}

We first study hybrid composition with IdleBlock on MobileNet v3. We evaluate different hybridizing configurations (Table \ref{tab: mbv3_hc}). With HC and $\alpha=0.5$ IdleBlock, if we seek to prune the network, we can prune up to 23\% of the MAdds with only approximately 0.7 percentage point Top-1 accuracy loss. If we seek similar MAdds, we can improve network accuracy by approximately 0.4 to 0.9 percentage point under different training settings. If we go deeper with HC, the deeper networks are potentially more efficient than all state-of-art human expert-designed network, and all state-of-art searched networks (Table \ref{tab: mbv3_all}).   

\subsection{Result on EfficientNet-B0}
\input{latex/tab/tab_efnet.tex}
\input{latex/tab/tab_efnet_all.tex}

Similar to the MobileNet-v3 experiment, we experiment with different configurations of hybrid composition on EfficientNet-B0 (Table \ref{tab: efb0_hc}).  With HC and $\alpha=0.5$ IdleBlock, if we seek to prune the network, we can prune up to 22\% of the MAdds with only approx 0.7 percentage point Top-1 accuracy loss, which is similar to observations on MobileNet v3. If we seek similar MAdds, we can improve network accuracy by approx 0.6 percentage point. If we go deeper with HC, the deeper networks are consistently more efficient. (Table \ref{tab: mbv3_all}).

\subsection{Inference Result on x86/ARM CPU}

\begin{figure}
\centering
\includegraphics[scale=0.35]{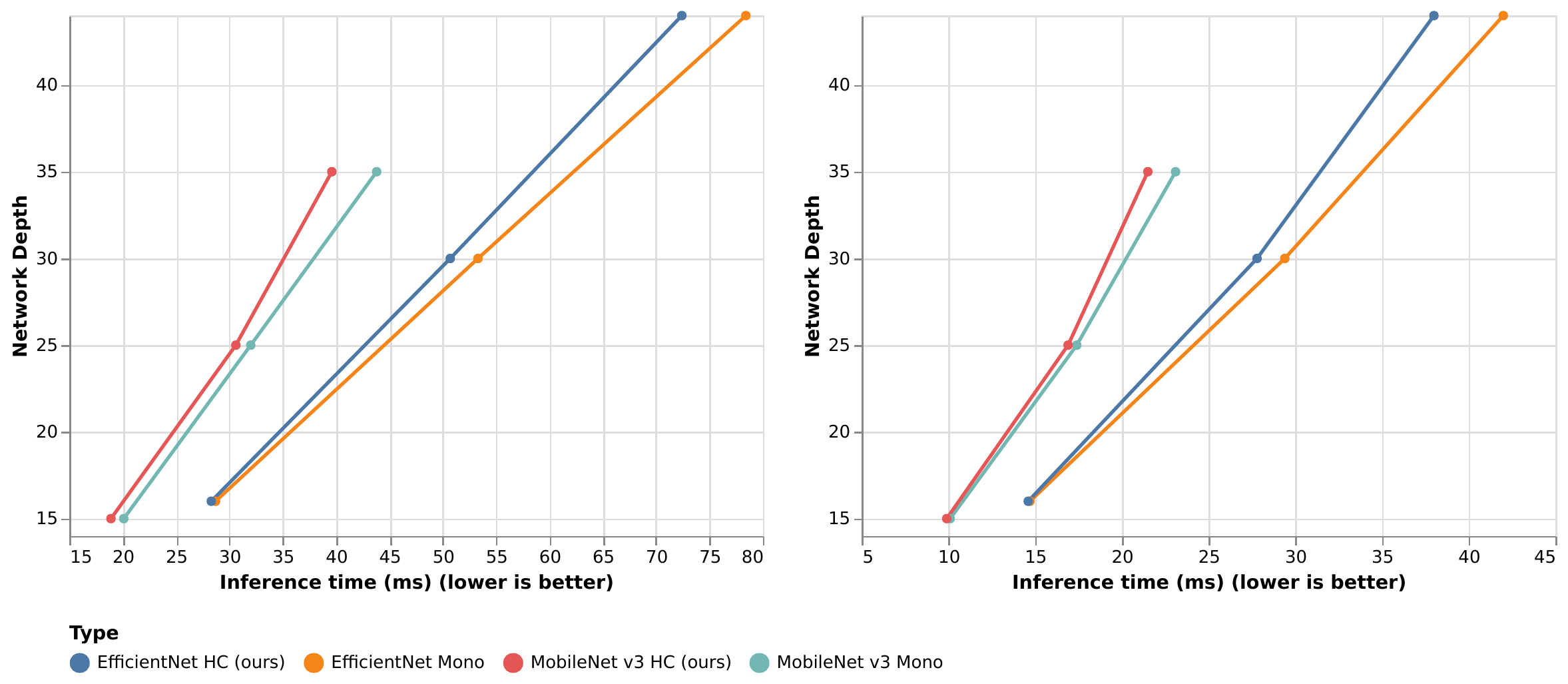}
\caption{x86 Inference on network with different depth.Left: 1 Thread; Right: 4 Threads}
\label{fig:x86}
\end{figure}

\begin{figure}
\centering
\includegraphics[scale=0.35]{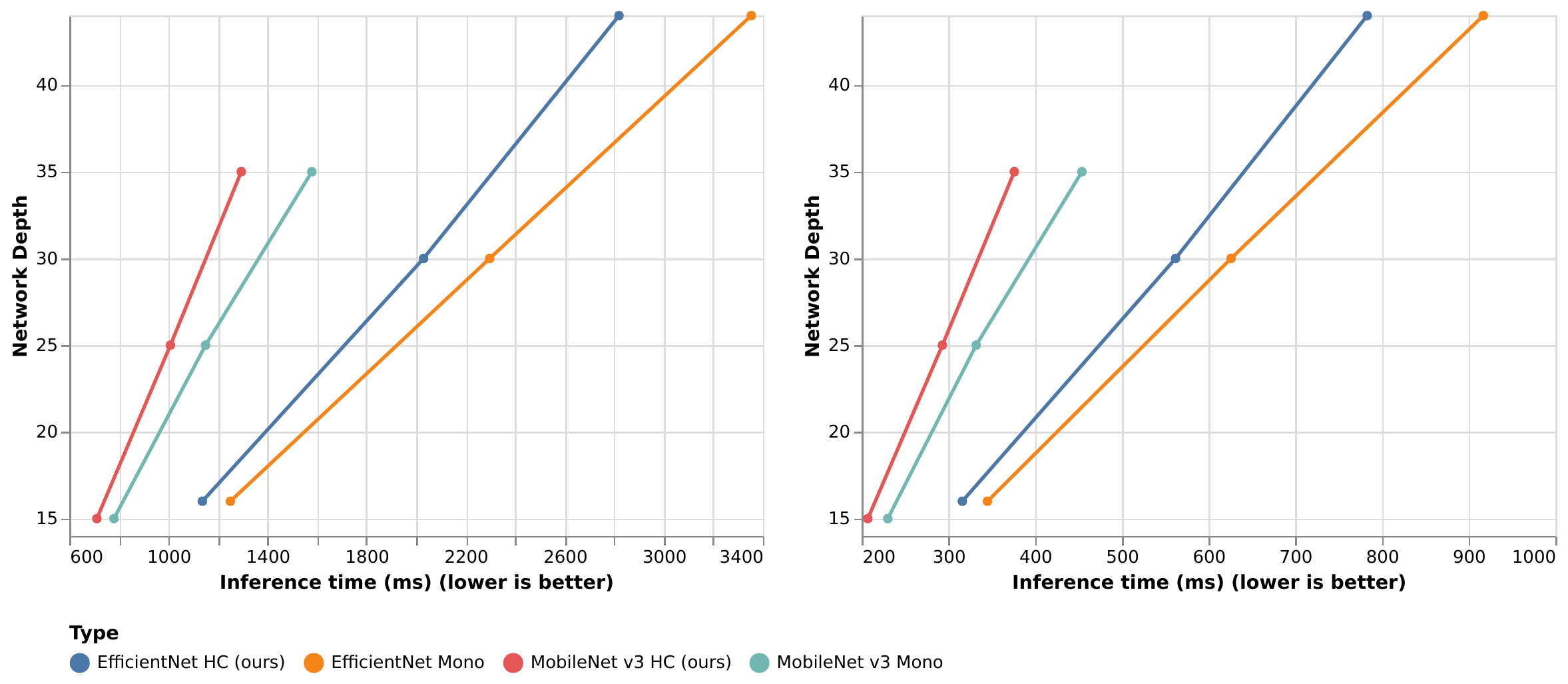}
\caption{ARM Inference on network with different depth. Left: 1 Thread; Right: 4 Threads}
\label{fig:arm}
\end{figure}

Our current implementation, based on TVM, realizes approximately one-third of the theoretical speedups obtained from our new block structure across a range of hardware. Future work may involve representing the entire block structure in the tensor expression IR, which will allow more aggressive optimizations such as completely eliminating the concatenation, splitting, and various layout transformation operations that are present in our implementation. Nevertheless, we still obtain real speed up with same depth without these specialized optimization on very different hardware platforms (Figure \ref{fig:x86}, Figure \ref{fig:arm}).

\subsection{Ablation Study}

\subsubsection{Impact of Channel Shuffle Operator}

In Section \ref{sec:idle} we mentioned that the channel shuffle operator is not friendly to heterogeneous  hardware accelerators for inference, despite its minimal theoretical computation complexity. In this experiment, we want to quantitatively understand channel shuffle operator’s contribution in network accuracy. We refer IdleBlock with channel shuffle operator as Inverted Shuffle Block (ISB). The difference between ISB and ShuffleBlock v2 is: ISB is based on MBBlock while ShuffleBlock v2 is based on a special case of Bottleneck block.

\begin{table}[t]
\centering
\small
\begin{tabular}{llll}
\toprule
Model                       & Madds & Params & Top-1 \\ \midrule
M=1, ISB=14                 & 157.8 & 4.8    & 73.8/73.7$^\star$      \\     
\textbf{M=1, Idle=14 }      & 157.8 & 4.8    & \textbf{73.8/73.8$^\star$}      \\  
\midrule
HC(M=10, ISB=10)            & 299.8 & 7.3    & 76.4/76.6$^\star$      \\  
\textbf{HC(M=10, Idle=10) } & 299.8 & 7.3    & \textbf{76.8/76.7$^\star$}      \\  
\midrule
HC(M=15. ISB=20)            & 380.1 & 8.1    & 77.0/77.2$^\star$      \\  
\textbf{HC(M=15, Idle=20) } & 380.1 & 8.1    & \textbf{77.2/77.4$^\star$}      \\ \bottomrule 
\end{tabular}

\caption{Ablation study on impact of channel shuffle operator. We adopt block property settings from MobileNet v3. ($^\star$) denotes number from distributed training.
}
\label{tab:isb}
\end{table}

We use MobileNet-v3 to study this problem. First we replace 14 out of 15 MBBlock to ISB or IdleBlock, only keep the very first MBBlock. We also tried different hybrid composition configurations in Table \ref{tab: mbv3_all}. Detailed results can be found in Table. \ref{tab:isb}. This experiment suggests that channel shuffle operators are unnecessary in our design.


\subsubsection{Deeper Net with Monotonous MBBlock vs Hybrid Composition with IdleBlock}

By applying hybrid composition with IdleBlock, we can save computational cost for a fixed depth. In this experiment, we want to understand the the accuracy difference between monotonous stacking with MBBlock vs hybrid composition with IdleBlock when going deeper. We study on both MobileNet v3 and EfficientNet-B0.

\begin{table}[t]
\small
\centering

\begin{tabular}{llll}
\toprule
MboileNet v3    & MAdds (M)  & Params (M) & Top-1 (\%) \\ \midrule
M=15, I=10      & 299.8  & 7.26   & 76.8/\textbf{77.0$^\star$}       \\
M=25, I=0       & 368.4  & 8.85   & 76.9/76.7$^\star$      \\ \hdashline
M=15, I=20      & 380.1  & 8.09   & 77.2/77.5$^\star$       \\
M=35, I=0       & 517.5  & 11.26  & \textbf{77.6}/77.2$^\star$       \\ \bottomrule
EfficientNet B0  & MAdds (M)  & Params (M) & Top-1 (\%) \\ \midrule
M=16, I=14      & 569.0  & 7.12   & 79.0       \\
M=30, I=0       & 715.94 & 8.32   & \textbf{79.2}       \\ \hdashline
M=16, I=28      & 752.2  & 9.18   & 79.5       \\
M=44, I=0       & 1046.0 & 11.57  & \textbf{79.8}       \\ \bottomrule
\end{tabular}

\caption{Ablation study on deep monotonous networks with MBBlock and deep hybrid composition networks with IdleBlock. In this table M indicates count of MBBlock, and I indicates count of IdleBlock. ($^\star$) denotes number from distributed training.}
\label{tab:deep}

\end{table}

From Table \ref{tab:deep}, we find stacking more MBBlock monotonically does not always guarantee better accuracy with different training methods. Stacking MBBlock usually costs 1.3X more MAdds to achieve the same depth, compared to hybrid composition with IdleBlock. The accuracy difference is approximately 0.3 percentage points at various depth settings.

\subsubsection{Impact of L/R-IdleBlock}


Empirically, for shallow networks, a larger receptive field will improve accuracy. We use MobileNet v3 to study this phenomena. Result is listed in Table \ref{tab:lr}.  We find L/R-IdleBlock placement has an impact on the final accuracy. Most of our experiments empirically validate our hypothesis, as monotonically stacked IdleBlock create a larger receptive field. In one of one settings, L/R cross placement performs better than monotonically use IdleBlock. L/R-Idle block placement introduce a new hyperparameter in network design, but this hyperparameter is substantially less sensitive than other hyper-parameters. 

\begin{table}[]
\centering

\begin{tabular}{lll}
\toprule
\small
MobileNet v3      & Top-1 & Dist. Top-1 \\ \midrule
M=1, L/R-Idle=14  & 73.09 & 73.48      \\  
M=1, Idle=14      & \textbf{73.82} & \textbf{73.78}      \\  \hdashline
M=5, L/R-Idle=10  & 74.23 & 74.42      \\  
M=5, Idle=10      & \textbf{74.67} & 74.42      \\  \hdashline
M=15, L/R-Idle=20 & 76.38 & \textbf{77.57}      \\  
M=15, Idle=20     & \textbf{77.20} & 77.39      \\ \bottomrule 
\end{tabular}

\caption{Ablation study on monotonically use IdleBlock or mix use L/R-IdleBlock with different depth settings under MobileNet v3. In this table M indicates number of MBBlock, L/R-IdleBlock indicates using L-IdleBlock and R-IdleBlock alternately.}
\label{tab:lr}
\end{table}

%% file: latex/tab/tab_mbv3.tex
\begin{table*}[t]
\centering
\small
\begin{tabular}{lllllllll}
\toprule
Config                   & MAdds & Params & \#MBBlock & \#IdleBlock & Top-1  & ARM & x86 \\
~                        & (M)       & (M)        & ~         & ~           & (\%)  & 1T/4T (ms) & 1T/4T (ms)\\
\hline
None (baseline)          & 219.4     & 6.44       & 15        & 0           & 75.20/75.11$^\star$ & 780/230 & 20.1/10.1                         \\ \hdashline
Maximum                    & \textbf{168.5}     & 5.51       & 5         & 10          & 74.67/74.42$^\star$    & 685/201 & 19.5/10.3                    \\ 
Adjacent                   & 185.9     & 5.64       & 10        & 5           & 75.03/74.86$^\star$   & 711/207 &18.9/9.9                         \\ 
Adjacent + 1 IdleBlock    & 224.4     & 6.06       & 10        & 10          &  75.58/75.82$^\star$   & 844/246 & 22.0/11.9                  \\ 
Adjacent + 1 IdleBlock L/R & 224.4     & 6.06       & 10        & 10          & \textbf{75.66/76.04$^\star$} &844/245 & 22.2/11.8                     \\
\bottomrule
\end{tabular}

\caption{Applying different Hybrid Composition configurations on MobileNet-v3. $\star$ indicates using distributed training. The “None” configuration is the standard MobileNet v3. “Adjacent + 1 IdleBlock L/R” is the configuration where we replace one MBBlock with one L-IdleBlock and one R-IdleBlock. When adding or substituting MBBlock with IdleBlock, we use the same SE \cite{SENet}, channels, and activation settings from the substituted MBBlock.}
\label{tab: mbv3_hc}
\end{table*}

%% file: latex/tab/tab_mbv3_all.tex
\begin{table}[t]
\centering
\begin{small}

\begin{tabular}{llll}
\toprule
Model              & MAdds (M) & Params (M) & Top-1 \\ \midrule
MobileNet v3 \cite{howard2019searching}       & 219       & 5.4        & 75.2     \\ 
FBNet-A \cite{wu2019fbnet}            & 249       & 4.3        & 73.0     \\ 
FBNet-B \cite{wu2019fbnet}            & 295       & 4.5        & 74.1     \\ 
MobileNet v2 \cite{sandler2018mobilenetv2}        & 300       & 3.4        & 72.0      \\
Proxyless \cite{cai2018proxylessnas}         & 320       & 4.1        & 74.6     \\

\midrule
FBNet-C \cite{wu2019fbnet}             & 375       & 5.5        & 74.9     \\ 
EfficientNet-B0 \cite{tan2019efficientnet}    & 390       & 5.3        & 76.3     \\ 
NASNet-B \cite{zoph2018learning}           & 488       & 5.3        & 72.5     \\ 
NASNet-C \cite{zoph2018learning}          & 558       & 4.9        & 74.5     \\ 
NASNet-A \cite{zoph2018learning}           & 564       & 5.3        & 74.0     \\ 
RandomWired-WS  \cite{xie2019exploring}    & 583       & 5.6        & 74.7     \\ 
AmoebaNet \cite{real2019regularized}           & 580       & 6.4        & 75.7     \\ 
MobileNet v2 (1.4) \cite{sandler2018mobilenetv2} & 585       & 6.9        & 74.7     \\ 
PNAS \cite{liu2018progressive}               & 588       & 5.1        & 74.2     \\ 
ShuffleNet v2 \cite{ma2018shufflenet}      & 591       & 7.4        & 74.9     \\ 
DARTS \cite{liu2018darts}              & 595       & 4.9        & 73.1     \\ 

\midrule
HC(M=5, I=10)      & 169       & 5.5        & 74.7     \\ 
HC(M=10, I=10)      & 224       & 6.0        & 75.7         \\ 
HC(M=15, I=10)     & 299       & 7.3        & 76.8     \\ 
HC(M=15, I=20)     & 380       & 8.1        & 77.2    \\ 
\midrule
HC(M=5, I=10)$^\star$       & 169       & 5.5        & 74.4     \\ 
HC(M=10, I=10)$^\star$      & 224       & 6.0        & 76.0         \\ 
HC(M=15, I=10)$^\star$      & 299       & 7.3        & 77.0     \\ 
HC(M=15, I=20)$^\star$      & 380       & 8.1        & 77.5    \\ 
\bottomrule
\end{tabular}

\end{small}

\caption{A comparison of Hybrid Composition with IdleBlock on MobileNet v3 with state-of-the-art human expert-designed \& NAS networks. Our work is listed as HC(M=x, I=y): M is total count of MBBlock and I is total count of IdleBlock. Models trained in a distributed setting are denoted by ($^\star$).}
\label{tab: mbv3_all}
\end{table}

%% file: latex/tab/tab_efnet.tex
\begin{table*}[t]
\centering
\small

\begin{tabular}{lllllllll}
\toprule
Config                   & MAdds & Params & \#MBBlock & \#IdleBlock & Top-1  & ARM & x86 \\
~                        & (M)       & (M)        & ~         & ~           & (\%)  & 1T/4T (ms) & 1T/4T (ms)\\
\hline
None                       & 385.3     & 5.07       & 16        & 0           & 77.05               & 1250/345 & 28.7/14.7              \\ \hdashline
Maximum                    & 299.7     & 4.21       & 7         & 9           & 76.34               & 1066/298 & 27.2/13.8                      \\ 
Adjacent                   & 334.4     & 4.54       & 11        & 5           & 76.61               & 1137/316 & 28.3/14.6                   \\ 
Adjacent + 1 IdleBlock    & 397.81    & 5.46       & 11        & 10          & \textbf{77.60 }      & 1373/384 & 32.2/17.2                             \\ 
Adjacent + 1 IdleBlock L/R & 397.81    & 5.46       & 11        & 10          & 77.25               & 1372/383 & 32.1/17.4                      \\ 
\bottomrule
\end{tabular}

\caption{Applying different Hybrid Composition configurations on EfficientNet-B0. As in the MobileNet v3 experiments, SE, channels, non-linear activations and DropConnect \cite{wan2013regularization} settings are set to be identical to the substituted MBBlock.}
\label{tab: efb0_hc}
\end{table*}

%% file: latex/tab/tab_efnet_all.tex
\begin{table}[t]

\centering

\begin{small}
\begin{tabular}{llll}
\toprule
Model            & MAdds (M) & Params (M) & Top-1 \\ \midrule
EfficientNet-B0 \cite{tan2019efficientnet}  & 390       & 5.3        & 76.3      \\ 
EfficientNet-B1 \cite{tan2019efficientnet}  & 700       & 7.8        & 78.8      \\ 
EfficientNet-B2 \cite{tan2019efficientnet}  & 1000      & 9.2        & 79.8      \\ 
EfficientNet-B3 \cite{tan2019efficientnet}  & 1800      & 12         & 81.1      \\ \hline
ResNet-50 v1$^\lozenge$     & 4100      & 26.0         & 77.3      \\ 
ResNeXt-50 32x4$^\lozenge$  & 4200      & 25.0         & 79.3      \\ 
ResNeXt-101 32x4$^\lozenge$  & 7800      & 44.3           & 80.3      \\ 
HRNet-W32-C \cite{sun2019deep}         & 8310      & 41.2       & 79.5      \\  
ResNet-152 v1$^\lozenge$    & 11000     & 60.0         & 79.2      \\
Squeeze-Excite-Net \cite{SENet}  & 21000     & 115.1        & 81.3      \\    
\midrule
EfficientNet-B0$^\star$ & 385       & 5.1        & 77.0      \\ 
\textbf{HC(M=16, I=14)}   & 569       & 7.1        & 79.0      \\ 
\textbf{HC(M=16, I=28) }  & 752       & 9.2        & 79.5          \\
HC(M=16, I=28)$^\square$ & 1532 & 9.2 & 81.0 \\

\bottomrule
\end{tabular}
\end{small}

\caption{A comparison of Efficient-B0 with hybrid composition to state-of-the-art models. ($^\star$) indicates our implementation. ($^\lozenge$) indicates results from GluonCV \cite{guo2019gluoncv}, which improves the results in the original paper due to the use of Mixup and other advanced training methods. ($^\square$) indicates the network is trained \& tested with images at $320 \times 320$ resolution.}
\label{tab:ef_all}

\end{table}

%% file: latex/6-conclusion.tex
In this paper, we introduce a new network design methodology inspired by network pruning: Idle and its MBBlock version IdleBlock. In order to utilize IdleBlock efficiently, we break the tradition of monotonous design in network stage and introduce hybrid composition of IdleBlock with MBBlock. \textbf{Our empirical results suggest that using the pruned computation to make the network deeper with hybrid composition is more efficient than various state-of-art neural architecture search methods for small models.} Hybrid composition also makes intermediate and large networks more efficient. This may point to a simpler direction for future neural architecture search directions for more efficient image recognition network. The principles of Idle design may also be worth to extending to neural networks for other tasks and domains.